\documentclass[sigconf]{acmart}

\AtBeginDocument{%
  }

\setcopyright{acmlicensed}
\copyrightyear{2018}
\acmYear{2018}
\acmDOI{XXXXXXX.XXXXXXX}
\acmConference[Conference acronym 'XX]{Make sure to enter the correct
  conference title from your rights confirmation email}{June 03--05,
  2018}{Woodstock, NY}
\acmISBN{978-1-4503-XXXX-X/2018/06}

\PassOptionsToPackage{table}{xcolor}
\usepackage{xcolor}
\usepackage{colortbl}
\usepackage{tabularx}       
\usepackage{multirow}
\usepackage{mathtools}
\usepackage{subfigure}
\usepackage{cleveref}
\usepackage{wrapfig}
\usepackage{graphicx}
\usepackage{scalerel}  

\definecolor{orange1}{HTML}{F4B183}
\definecolor{lightblue}{HTML}{E3EEF8}

\newcommand{\red}[1]{\textcolor{red}{#1}}

\newcommand{\bestimp}[2]{%
 \ensuremath{
    \textbf{#1}
    \mathrlap{{\textcolor{red}{\scalebox{0.75}{\textbf{+#2\%}}}}}
 }
}
\newcommand{\imp}[2]{%
 \ensuremath{
    #1
    \mathrlap{{\textcolor{red}{\scalebox{0.75}{\textbf{+#2\%}}}}}
 }
}

\begin{document}


\title[Self-Improving Unified LMMs with Dual Self-Rewards]{SUDER: Self-Improving Unified Large Multimodal Models for Understanding and Generation with Dual Self-Rewards}

\author{Jixiang Hong}
\affiliation{%
  \institution{Gaoling School of Artificial Intelligence, \\ Renmin University of China}
  \city{Beijing}
  \country{China}
}
\email{jxhong@ruc.edu.cn}

\author{Yiran Zhang}
\affiliation{%
  \institution{School of Computer Science and Technology, \\ UCAS}
  \city{Beijing}
  \country{China}}
\email{zhangyiran232@mails.cas.edu.cn}

\author{Guanzhong Wang}
\author{Yi Liu}
\affiliation{%
  \institution{Baidu Inc.}
  \city{Beijing}
  \country{China}}
\email{wangguanzhong@baidu.com}
\email{liuyi22@baidu.com}

\author{Ji-Rong Wen}
\affiliation{%
  \institution{Gaoling School of Artificial Intelligence, \\ Renmin University of China}
  \city{Beijing}
  \country{China}
}
\email{jrwen@ruc.edu.cn}

\author{Rui Yan}\authornote{Corresponding author: Rui Yan(ruiyan@ruc.edu.cn)}
\affiliation{%
  \institution{ Gaoling School of Artificial Intelligence, \\ Renmin University of China}
  \city{Beijing}
  \country{China}
}
\affiliation{%
  \institution{School of Computer Science, Wuhan University}
  \city{Wuhan}
  \country{China}}
\email{ruiyan@ruc.edu.cn}

\renewcommand{\shortauthors}{Hong et al.}

\begin{abstract}

Building upon large language models (LLMs), recent large multimodal models (LMMs) unify cross-model understanding and generation into a single framework.
However, LMMs still struggle to achieve accurate vision-language alignment, prone to generating text responses contradicting the visual input or failing to follow the text-to-image prompts.
Current solutions require external supervision (e.g., human feedback or reward models) and only address unidirectional tasks—either understanding or generation.
In this work, based on the observation that understanding and generation are naturally inverse dual tasks, we propose \textbf{SUDER} (\textbf{S}elf-improving \textbf{U}nified LMMs with \textbf{D}ual s\textbf{E}lf-\textbf{R}ewards), a framework reinforcing the understanding and generation capabilities of LMMs with a self-supervised dual reward mechanism. 
SUDER leverages the inherent duality between understanding and generation tasks to provide self-supervised optimization signals for each other.
Specifically, we sample multiple outputs for a given input in one task domain, then reverse the input-output pairs to compute the dual likelihood within the model as self-rewards for optimization.
Extensive experimental results on visual understanding and generation benchmarks demonstrate that our method can effectively enhance the performance of the model without any external supervision, especially achieving remarkable improvements in text-to-image tasks.

\end{abstract}



\received{20 February 2007}
\received[revised]{12 March 2009}
\received[accepted]{5 June 2009}

\maketitle

\section{Introduction}
Recently, following the success of LLMs in text-centric tasks, many works have explored extending LLMs to multimodal tasks, leading to the emergence of LMMs.
By optionally incorporating modality-specific encoders and decoders (or tokenizers and de-tokenizers), some LMMs are tailored for visual understanding tasks~\citep{bai2023qwen,liu2023visual}, some for visual generation tasks~\citep{sun2024autoregressive}, while others are designed to unify both understanding and generation tasks~\citep{wang2024emu3,xie2024show,wu2024janus,chen2025janus}.

\begin{figure}[t]
    \centering
    \subfigure[Misalignment in visual understanding.]{
        \label{fig:example1}
        \includegraphics[width=0.8\linewidth]{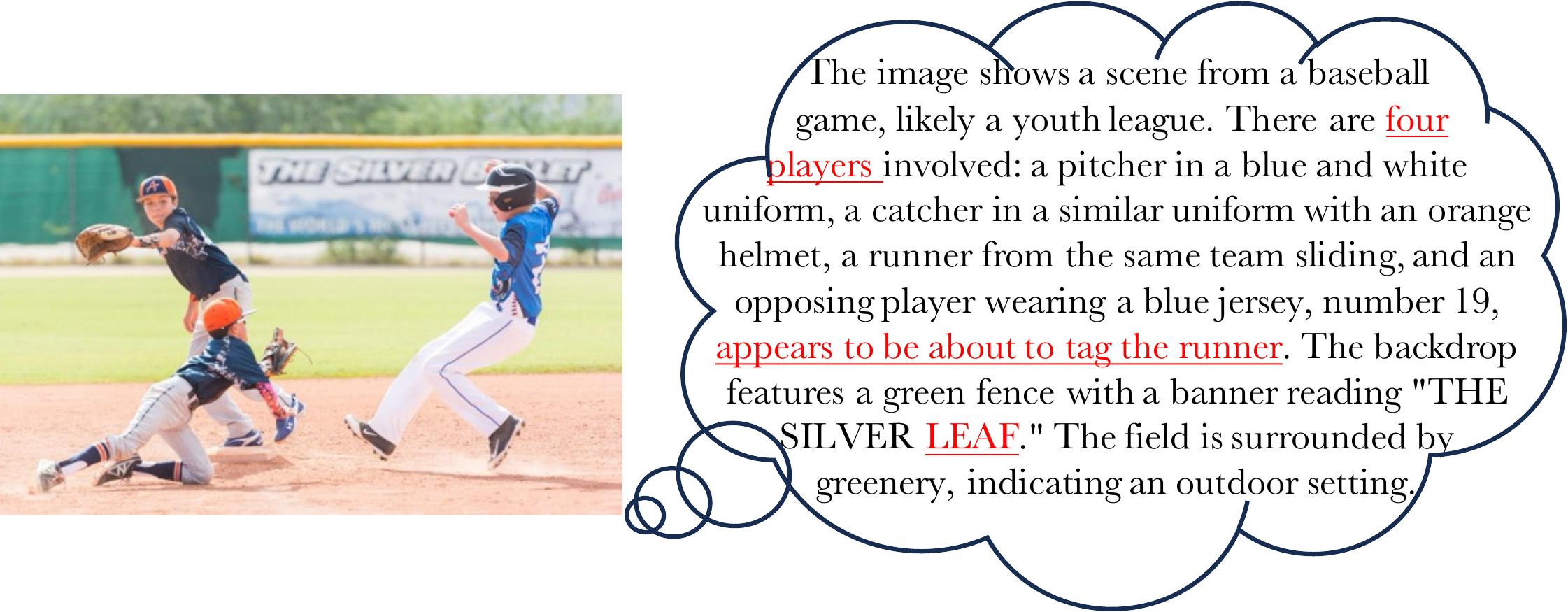}
    }
    \subfigure[Misalignment in text-to-image generation.]{
        \label{fig:example2}
        \includegraphics[width=0.8\linewidth]{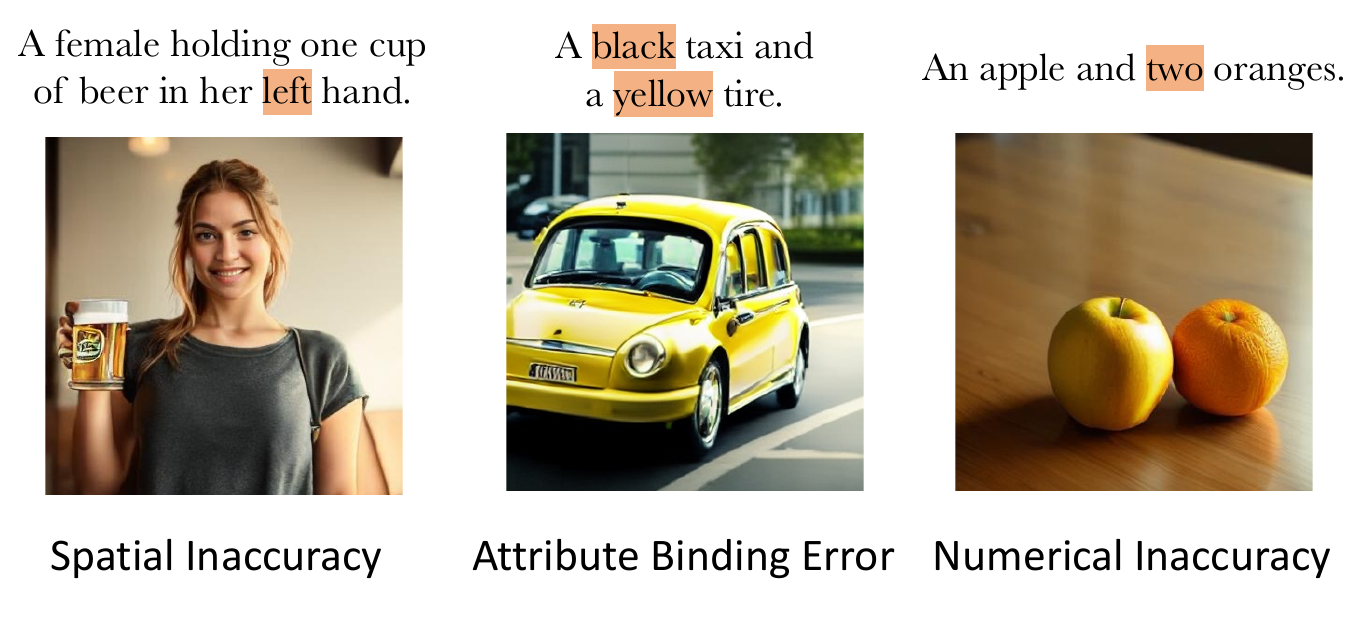}
    }
    \caption{Failures in visual understanding and image generation of Janus-Pro-7B. In (a), the text in \red{$\underline{\text{red}}$} indicates error content. In (b), the generated images do not follow the content in \colorbox{orange1}{orange}.}
    \label{fig:example}
\end{figure}

Despite the impressive performance of LMMs in both understanding and generation tasks, they still face challenges in aligning text and image modalities. 
For visual understanding tasks, LMMs often fail to accurately perceive visual content and return hallucinated or incorrect responses, as illustrated in \Cref{fig:example1}.
\citet{ji2024align,ouali2024clip} conduct supervised fine-tuning (SFT) on paired data with preference labels to mitigate this problem.
For text-to-image generation tasks, LMMs tend to generate images that are inconsistent with the text prompts, as shown in \Cref{fig:example2}.
\citet{jiang2025t2ir1} propose a two-level chain-of-thought (COT) generation method and employ ensemble rule-based reward models to optimize the model.
\citet{wang2025simplear} utilize CLIP~\citep{pmlr-v139-radford21a} as the reward model to supervise the reinforcement learning of generation.
These efforts often concentrate on improving only one aspect (either understanding or generation) of LMM capabilities, and most of them depend heavily on external supervision signals or parallel text-image pairs data for training.

To address these limitations, we explore how to reinforce the understanding and generation of LMMs in a self-supervised manner, without relying on external supervision and pairing data.
We observe that understanding and generation are inherently dual tasks, where the output of one task can naturally serve as the input of the other. 
Moreover, existing unified LMMs have already undergone basic alignment and demonstrate a certain level of capability in both tasks, but the dual-task relationship is not fully utilized.
Motivated by this, we seek to exploit this duality between understanding and generation to enable mutual self-supervision, allowing the optimization of one task to be guided by feedback derived from its dual counterpart.
By this, we can achieve self-improvement of the unified LMMs in both understanding and generation tasks without the need for external human or AI supervision.
However, to derive self-supervised rewards for both tasks in a unified and efficient way is non-trivial. 
Existing methods tend to employ visual query-answer (VQA) tasks to assess the generated images in text-to-image (T2I) tasks~\citep{qu2025silmm,wang2024illume}, which are not applicable to understanding tasks. 
DeGF~\citep{zhang2025selfcorrecting} leverages an auxiliary image generated by Stable Diffusion~\citep{rombach2022high} to improve the understanding performance, which relies on feedback from external models and is of high cost and latency. 

To achieve our objective and tackle the above challenges, we propose a dual self-reward mechanism to provide self supervision for optimizing the understanding and generation of LMMs.
Specifically, for visual understanding, we sample multiple textual descriptions given an input image, then reverse the input-output pairs and use each description as the condition to compute the likelihood of generating the original image. This enables a quantitative assessment of how well each description aligns with the visual content.
Analogously, for text-to-image generation, we compute the likelihood of generating the original text prompt conditioned on the sampled images to assess their semantic fidelity.
By incorporating this dual self-reward mechanism, we introduce a framework self-improving unified LMMs SUDER (\textbf{S}elf-improving \textbf{U}nified LMMs with \textbf{D}ual s\textbf{E}lf-\textbf{R}ewards), which can effectively optimize the understanding and generation capabilities in a unified manner.
Besides, we explore different optimization strategies, including jointly optimizing both capabilities within a single model, as well as alternately optimizing the two capabilities in two adversarial models. 
We also investigate different optimization algorithms for this task, including preference optimization and reinforcement learning algorithms.

Extensive experiments have been conducted on various visual understanding and generation benchmarks to validate the effectiveness of our method. 
Specifically, we evaluate the performance of our method on the T2I-CompBench~\citep{huang2023t2i} and GenEval~\citep{ghosh2023geneval} datasets, which are widely used for assessing the performance of LMMs in text-to-image generation tasks. 
For example, our tuned Janus-Pro-7B~\citep{chen2025janus} model gains remarkable improvements over the baseline model, achieving an 11.68\% increase on average on T2I-CompBench and a 5\% increase on GenEval.
Additionally, the model also demonstrates improved performance on visual understanding tasks, for example, achieving a 7.5\% increase in the overall score on the LLaVA-Bench~\footnote{https://huggingface.co/datasets/liuhaotian/llava-bench-in-the-wild}.

Overall, our contributions can be summarized as follows:

\begin{itemize}
    \item We present SUDER, a unified self-supervised framework that jointly enhances both the understanding and generation capabilities of large multimodal models (LMMs).
    \item We propose a dual self-reward mechanism that leverages the inherent duality between understanding and generation tasks to provide self-supervised optimization signals. And we explore different training strategies and optimization algorithms based on the dual self-rewards for unified learning of LMMs.
    \item We conduct extensive experiments on various visual understanding and generation benchmarks, demonstrating the effectiveness of our method in improving the performance of unified LMMs.
\end{itemize}

\section{Related Work}

\subsection{Multimodal Understanding and Generation}

\paragraph{Visual Understanding.} Advancement in large language models (LLMs) has paved the way for the development of large multimodal models (LMMs) for vision-language understanding. 
These works focus on how to align visual signals with the language semantic space of the pre-trained LLMs.
By connecting vision encoders to pre-trained LLMs, LMMs can effectively understand both visual and textual inputs~\citep{openai2023gpt4v, anthropic2024claude}.
LLaVA~\citep{liu2023visual} and MiniGPT-4~\citep{zhu2024minigpt} were among the first to attempt to train adapters to shift the output of the vision encoder into the embedding space of the language model.
Qwen-VL~\citep{bai2023qwen}, InternVL~\citep{chen2024intern} and Deepseek-VL~\citep{lu2024deepseek, wu2024deepseek} further enhance the visual-linguistic performance on more complex scenarios by incorporating more advanced LLM backbones, additional training data of higher quality and more effective training strategy. 
Although these models have achieved impressive performance in visual understanding and reasoning, mere perception and comprehension fall short of our broader expectations for multimodal intelligence.

\paragraph{Visual Generation.} 

Recent advances in visual generation can be generally categorized into two main approaches: diffusion models and autoregressive models. 
The LMMs for visual generation fall in the latter category, which we mainly focus on in this paper. 
These models typically adopt an autoregressive approach, which is the focus of this paper. These models generate images by producing a sequence of discrete visual tokens, which are then decoded into images using vector-quantized (VQ) model~\citep{van2017neural,esser2021taming} based visual tokenizers. 
Representative works in this line of research include LlamaGen and DeLVM, which demonstrate the potential of autoregressive modeling in high-fidelity image synthesis within a multimodal framework.

\paragraph{Unified LMM for Understanding and Generation.}

Unified LMMs aim to integrate both understanding and generation capabilities into a single framework, enabling them to perform a wide range of multimodal tasks.
Unified LMMs aim to seamlessly integrate understanding and generation capabilities within a single framework, enabling them to handle a broad spectrum of multimodal tasks. 
Models like Emu~\citep{sun2023emu}, Emu2~\citep{sun2024generative}, X-ViLa~\citep{ye2024x}, and Next-GPT~\citep{wu2024next} adopt a unified autoregressive architecture that treats both visual embeddings and text tokens as elements to predict. 
AnyGPT~\citep{zhan2024anygpt}, Show-o~\citep{xie2024show}, ViLA-U~\citep{wu2024vila}, Chameleon~\citep{lu2023chameleon}, Emu3\citep{wang2024emu3}, and Janus series~\citep{wu2024janus,chen2025janus}, leverage VQ tokenizers to discretize images into tokens, thereby allowing both vision and language to be modeled uniformly through next-token prediction. 

\subsection{Optimization for Large Multimodal Models}
Although the LMMs demonstrated promising performance in various multimodal tasks, they still face challenges in accurately aligning textual and visual modalities. 
Existing works explore how to conduct fine-grained optimization on LMMs to further improve them.
Align-anything~\citep{ji2024align} propose a framework to align all-modality (any-to-any) models with human intentions using unified language feedback and 200k human preference annotations.
SILMM~\citep{qu2025silmm} proposes a self-VQA manner and DPO~\citep{rafailov2023direct} to improve the compositional text-to-image generation performance, while T2I-R1 utilizes ensemble rule-based reward models to reinforce two levels of chain-of-thought (COT) text-to-image generation with GRPO~\citep{shao2024deepseekmath}.
These methods focus on improving the performance of LMMs in either understanding or generation tasks, and most of them rely on external supervision signals, such as human feedback or reward models, which may not be readily available in all scenarios.
Based on the observation that understanding and generation are inherently dual tasks and they can assess the output of each other, we explore a self-supervised dual reward mechanism to reinforce the bi-directional tasks in a unified manner.

\begin{figure*}[t]
    \centering
    \includegraphics[width=0.9\textwidth]{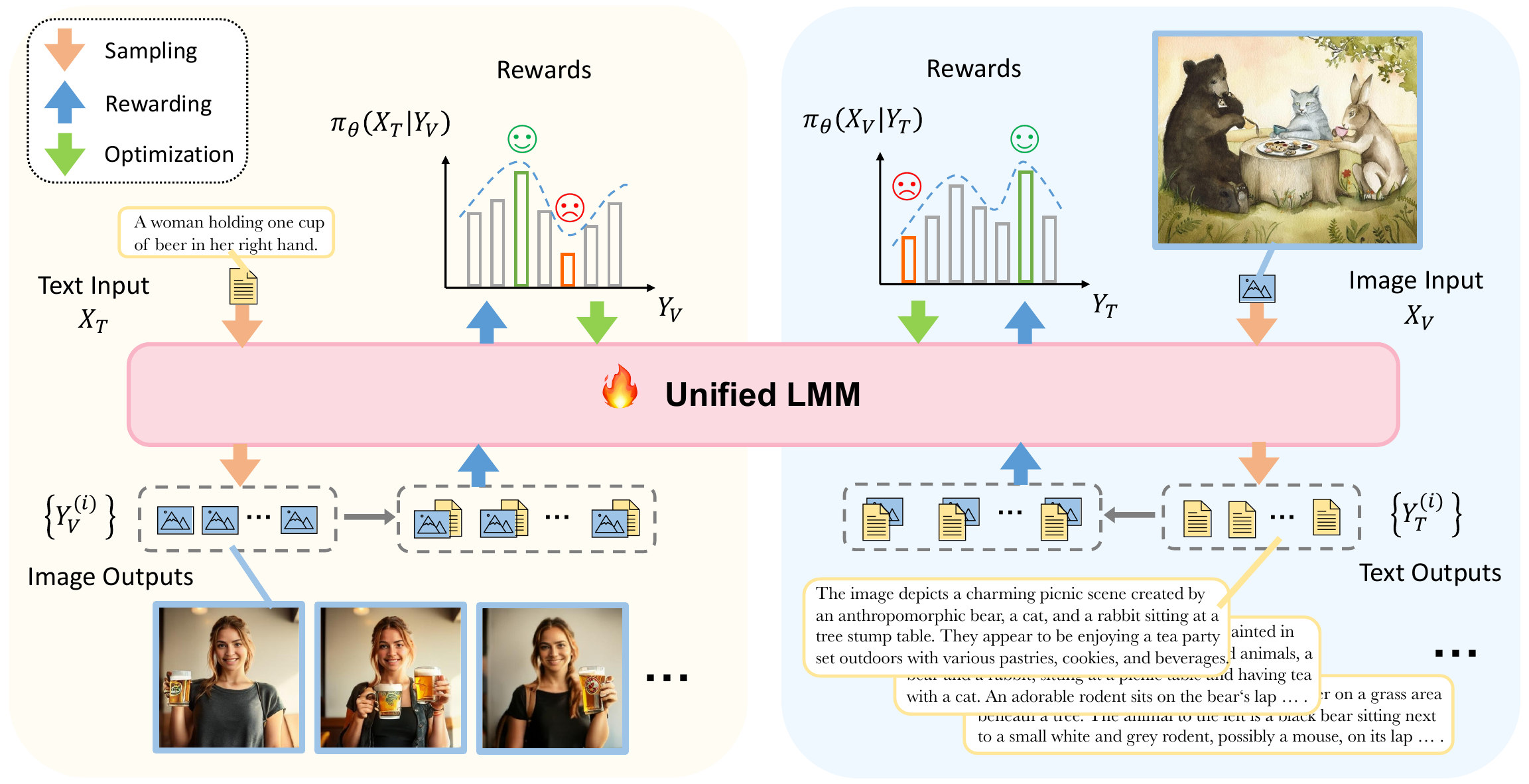}
    \caption{Overview of SUDER. The model samples multiple outputs for a given input for one task and reverses the input-output pairs to compute the dual likelihoods of the model outputs conditioned on the original input as self-rewards, which are utilized in optimization algorithms.}
    \label{fig:pipeline}
\end{figure*}

\section{Methodology}

In this section, we first introduce the duality between understanding and generation tasks in unified LMMs. 
Then we present our dual self-reward mechanism, which leverages the inherent duality between understanding and generation tasks to provide self-supervised optimization signals. 
Finally, we describe the framework SUDER, which incorporates the dual self-reward mechanism and explores different optimization strategies and algorithms for unified learning of LMMs.
The overview of SUDER is shown in~\Cref{fig:pipeline}.

\subsection{Duality between Understanding and Generation}
\label{sec:duality}
The understanding and generation tasks in unified LMMs are inherently dual tasks. 
For text-to-image generation, the model takes a text prompt $X_{T}$ describing the desired content of the image as input and generates an image $Y_{V} = \{y_{v}^{i}\}_{i=1}^{\|Y_V\|}$ that aligns with the text.
This process can be formulated as:
\begin{equation}
\pi_{\theta}(Y_V \mid X_T) = \prod_{i=1}^{\|Y_V\|} \pi_{\theta}(y_{v}^{i} \mid X_T, y_{v}^{<i}),
\end{equation}
where each vision token $y_{v}^{i}$ is generated sequentially based on the previous tokens $y_{v}^{< i}$ and the text prompt $X_{T}$, and $\pi_{\theta}$ denotes the generation process parameterized by $\theta$. $\|Y_V\|$ is the number of vision tokens in the generated image.

Reversely, for visual understanding, we take the captioning task as the proxy.
The model takes an image $X_{V}$ as input and generates a text response $Y_{T} = \{y_{t}^{i}\}_{i=1}^{\|Y_{T}\|}$ that matches the content of the image. This process can be formulated as:
\begin{equation}
\pi_{\theta}(Y_T \mid X_V) = \prod_{i=1}^{\|Y_{T}\|} \pi_{\theta}(y_{t}^{i} \mid X_V, y_{t}^{<i}),
\end{equation}
where $y_{t}^{i}$ denotes the text tokens, and $\|Y_{T}\|$ is the number of text tokens in the generated response.

These two tasks exhibit natural duality, where the output of one task can serve as the input for the other, especially text-to-image generation and image captioning.

\subsection{Dual Self-reward Mechanism}
\label{sec:dual_self_reward}
Based on the duality between visual understanding and generation in LMMs, we propose a dual self-reward (DSR) mechanism to provide self-supervised optimization signals for both tasks.

Specifically, for visual understanding, we generate multiple textual descriptions for a given input image; and then reverse the input-output position by treating each generated description as the conditional input and the original image as the target output to compute the likelihood of the image conditioned on each textual description.
This procedure is based on the intuitive assumption that: 
if a textual description accurately reflects the visual input, then the likelihood of generating the original image conditioned on this description should be high. 
By this, we can effectively quantify how well each description explains and matches the visual content.
Similarly, for text-to-image generation, we sample multiple images given a text prompt and compute the likelihood of generating the original text prompt conditioned on each sampled image, which allows us to assess the semantic fidelity of the generated images.
We derive the dual self-rewards from the aforementioned reversed likelihoods, which can be expressed as:
\begin{equation}
\begin{aligned}
    R_{U}(Y_T \mid X_V) &= \frac{1}{\|X_{V}\|} \log \pi_{\theta}(X_{V}\mid Y_{T}) \\
    &= \frac{1}{\|X_{V}\|} \sum_{1}^{\|X_{V}\|} \log \pi_{\theta}(x_v^{i} \mid Y_{T}, x_{v}^{<i}),  \\
    R_{G}(Y_V \mid X_T) &= \frac{1}{\|X_{T}\|} \log \pi_{\theta}(X_{T}\mid Y_{V}) \\
    &= \frac{1}{\|X_{T}\|} \sum_{1}^{\|X_{T}\|} \log \pi_{\theta}(x_t^{i} \mid Y_{V}, x_{t}^{<i}),
\end{aligned}
\end{equation}
where $R_{U}(Y_T \mid X_V)$ and $R_{G}(Y_V \mid X_T)$ denote the self-rewards for textual response $Y_T$ given visual input $X_V$  in visual understanding task and visual output $Y_V$ given textual prompt $X_T$ in T2I generation tasks, respectively.

Our dual self-reward mechanism naturally possesses the following advantages: 
1) efficiency, compared to self-VQA and generating additional reference content, dual self-rewards can be computed in a single forward pass of the model; 
2) self-supervised, the dual self-rewards are derived from the model's own outputs, eliminating the need for external supervision or additional data;
3) unified, the dual self-rewards can be applied to both understanding and generation tasks, allowing for a more comprehensive optimization process;
4) unbiased, the rewards are likelihoods of the same original input for multiple sampled outputs, which can reduce the length bias~\citep{shen2023loose} existing in the reward model based methods~\citep{ji2024align}.

\subsection{SUDER: Self-improving ULMMs with DSR}
SUDER is a framework that incorporates the dual self-reward mechanism to the optimization process.
With the dual self-rewards, we can optimize the model using different optimization algorithms, including preference optimization and reinforcement learning algorithms, as well as different training strategies.

We adopt the online SimPO~\citep{meng2024simpo} method, which is a simple preference optimization method that can be easily applied to LMMs. 
We also employ the GRPO~\citep{shao2024deepseekmath} method, which is a prevalent reinforcement learning method to optimize the reasoning capabilities of LLMs.

Specifically, for each task (visual understanding or generation), we sample \(G\) candidate outputs and compute their dual self-rewards.
Then we select the highest-reward sample \( Y^{+} \) and the lowest-reward sample \( Y^{-} \), and train the model with a pairwise preference objective that encourages the model to prefer the better output.
The SimPO objective is defined as:
\begin{align}
\mathcal{J}_{\text{SimPO}}(\theta) 
&= \mathbb{E}_{X \sim \mathcal{D}, (Y^{+}, Y^{-}) \sim \pi_{\theta}(\cdot \mid X)} \notag \\
&\quad \left[ \log \sigma \left( \frac{\beta}{\|Y^{+}\|} \log \pi_{\theta}(Y^{+} \mid X) \right. \right. \notag \\
&\quad \left. \left. - \frac{\beta}{\|Y^{-}\|} \log \pi_{\theta}(Y^{-} \mid X) - \gamma \right) \right].
\end{align}
where \( \beta \) is a scaling factor for the preference margin, and \( \gamma \) is a reward margin used to enforce a separation between the better and worse outputs.
where \( R(\cdot \mid \cdot) \) denotes the corresponding dual self-reward from Section~\ref{sec:dual_self_reward}, and \( \sigma(\cdot) \) is the sigmoid function.
This objective encourages the model to assign a higher likelihood to outputs that better align with the input according to the DSR.

We also employ the GRPO~\citep{shao2024deepseekmath} method, which estimates the advantage in a group-relative manner without relying on a value function, making it suitable for optimizing LMMs with self-reward signals.
Following DAPO~\citep{yu2025dapo}, we adopt a higher clipping threshold and a token-level policy gradient loss in the GRPO objective. 

Given an input \( X \) (either \( X_T \) for generation or \( X_V \) for understanding), the model samples a group of \( G \) outputs \( \{Y^{(i)}\}_{i=1}^{G} \) under the reference policy \( \pi_{\theta_{\text{ref}}} \), each associated with a dual self-reward \( R^{(i)} = R(Y^{(i)} \mid X) \) as defined in Section~\ref{sec:dual_self_reward}.

The group-normalized advantage for token \( t \) in the \( i \)-th output is defined as:
\begin{equation}
    \hat{A}_{i,t} = \frac{R^{(i)} - \text{mean}(\{R^{(j)}\}_{j=1}^{G})}{\text{std}(\{R^{(j)}\}_{j=1}^{G})}.
\end{equation}

The token-level importance weight is defined as:
\begin{equation}
    r_{i,t}(\theta) = \frac{\pi_{\theta}(y^{(i)}_t \mid X, y^{(i)}_{<t})}{\pi_{\theta_{\text{ref}}}(y^{(i)}_t \mid X, y^{(i)}_{<t})}.
\end{equation}

Then the GRPO objective with KL regularization and clipping is then given by:
\begin{align}
&\mathcal{J}_{\text{GRPO}}(\theta) = \mathbb{E}_{X \sim \mathcal{D}, \{Y^{(i)}\}_{i=1}^{G} \sim \pi_{\theta}(\cdot \mid X)} \notag \\
&\left[
\frac{1}{\sum_{i=1}^{G} \|Y^{(i)}\|} \sum_{i=1}^{G} \sum_{t=1}^{\|Y^{(i)}\|}
\left(
\texttt{minclip}(r_{i,t}(\theta)\hat{A}_{i,t})
- \beta D_{\text{KL}}(\pi_{\theta} \parallel \pi_{\text{ref}})
\right)
\right], \notag \\
&\texttt{minclip}(r_{i,t}(\theta)\hat{A}_{i,t}) =
\min \left(
\begin{aligned}
& r_{i,t}(\theta)\hat{A}_{i,t}, \\
& \text{clip}(r_{i,t}(\theta), 1-\epsilon_{\text{low}}, 1+\epsilon_{\text{high}})\hat{A}_{i,t}
\end{aligned}
\right),
\end{align}
where \( \|Y^{(i)}\| \) is the length of the sampled output, \( \epsilon_{\text{low}} \) and \( \epsilon_{\text{high}} \) are the clipping thresholds, and \( \beta \) controls the KL divergence penalty with respect to a reference policy \( \pi_{\text{ref}} \).
This objective encourages the model to improve outputs with relatively high rewards while staying close to the original behavior policy and preventing overfitting or reward hacking.

Besides, we also design different optimization strategies for the two tasks, including jointly optimizing both capabilities within a single model, as well as alternately optimizing the two capabilities in two adversarial models.

\section{Experiments}
\label{sec:experiments}

\subsection{Experimental Setup}

\paragraph{Training Data}
We used the training set of T2I-CompBench~\citep{huang2023t2i} for the text-to-image generation part, which contains 5,600 text prompts. 
For the visual understanding part, we randomly sampled 2,800 images from the JourneyDB~\citep{sun2023journeydb} and COCO118K~\citep{li2024llava} derived from COCO Caption~\citep{lin2014microsoft} respectively.
It is worth noting that the image and text training data are non-parallel, meaning that we did not use any annotated image-text pairs to train the model. 

\paragraph{Baselines}
We employ the unified LMMs with different levels of capabilities as our backbones, including Janus-Pro-7B and Janus-Pro-1B~\citep{chen2025janus}, optimizing them with DSR in a self-supervised manner.
We compare our tuned model with state-of-the-art (SOTA) models on visual understanding and generation tasks, including models that are designed only for understanding such as LLaVA~\citep{liu2023visual} and InstructBLIP-7B~\citep{dai2023instructblip}, 
models designed only for generation tasks like CoMat~\citep{jiang2024comat} and LDM~\citep{rombach2022high}, as well as unified LMMs that can perform both tasks such as Chameleon~\citep{lu2023chameleon} and Show-o~\citep{xie2024show}.
All baselines can be found in \Cref{tab:t2i_compbench}, \Cref{tab:geneval}, and \Cref{tab:und_bench}.


\paragraph{Evaluation Benchmarks}
We evaluate our method on visual understanding and generation benchmarks and follow their
default settings. 
For visual generation, we use T2I-CompBench~\citep{huang2023t2i} and GenEval~\citep{ghosh2023geneval} as the evaluation benchmarks. 
We test on their original text prompts without any modification.
T2I-CompBench comprises 6,000 compositional text prompts evaluating 3
categories (attribute binding, object relationships, and complex compositions) and 6 sub-categories (color binding, shape binding, texture binding, spatial relationships, non-spatial relationships, and complex compositions). 
GenEval~\citep{ghosh2023geneval} includes 6 tasks (single object, two objects, counting, colors, position, and color attributes).
For visual understanding, we use HallusionBench (denoted as HalluBench)~\citep{zhao2023beyond}, LLaVABench, POPE~\citep{li2023evaluating}, MMB~\citep{liu2024mmbench}, and SEEDBench-IMG (denoted as SEED)~\citep{li2024seed}.

\subsection{Results on Visual Generation}
\label{sec:results_gen}

We compare our models with the SOTA diffusion models and unified LMMs on the T2I-CompBench and GenEval benchmarks. The results are shown in \Cref{tab:t2i_compbench} and \Cref{tab:geneval}, respectively.
Our trained models are denoted as ``Janus-Pro-7B + SUDER'' and ``Janus-Pro-1B + SUDER'', where the models are optimized with DSR and SimPO in the unified strategy.

The results show that our models achieve substantial improvements over the baseline models. 
For Janus-Pro-7B, there is an average increase of 11.68\% on T2I-CompBench and overall 5\% on GenEval, while Janus-Pro-1B achieves an average increase of 10.5\% on T2I-CompBench and 4\% on GenEval. 
On T2I-CompBench, our tuned models remarkably improve the performance in all categories, especially in attribute binding with an average improvement of 20\% and 36\%, respectively, for Janus-Pro-7B and Janus-Pro-1B.
In spatial relationships and complex compositions, our tuned Janus-Pro-7B also achieve impressive increases of 5\% and 3\% respectively, and Janus-Pro-1B achieves 17\% and 10\% respectively.
Our tuned model demonstrates the best performance in shape binding, texture binding, and complex compositions, and achieves the second best performance in color binding.
On GenEval, our models improve over the baseline model, especially in counting and color attributes with an increase of 13\% and 7\% respectively for Janus-Pro-7B, and in counting and and position with an increase of 6\% and 12\% respectively for Janus-Pro-1B.
The larger one outperforms the diffusion models overall, achieving the best performance.
These results demonstrate the effectiveness and superiority of our method in improving the text-to-image generation capabilities of LMMs, despite the fact that our model is optimized in a completely self-supervised manner without any external supervision or additional data.

\begin{table*}[ht]
\centering
\caption{Evaluation on T2I-CompBench. Und. and Gen. denote ``understanding'' and ``generation'', respectively. Higher ($\uparrow$) values indicate better performance. The best score is in \textbf{bold}, with the second best score \underline{underlined}. Line in \colorbox{lightblue}{blue} is our model.}
\label{tab:t2i_compbench}
\setlength{\tabcolsep}{0pt}
\begin{tabular}{lcccccc}
\toprule
\multicolumn{1}{c}
{\multirow{2}{*}{\textbf{Model}}} & \multicolumn{3}{c}{\textbf{Attribute Binding}} & \multicolumn{2}{c}{\textbf{Object Relationship}} & \multirow{2}{*}
{\textbf{\;Complex$\uparrow$\;}} \\
\cmidrule(lr
){2-4}
\cmidrule(lr){5-6}
& \textbf{ Color$\uparrow$} & \textbf{Shape$\uparrow$} & \textbf{Texture$\uparrow$} & \textbf{Spatial$\uparrow$} & \textbf{Non-Spatial$\uparrow$} & \\
\midrule
\multicolumn{7}{c}{\textit{Gen. Only}} \\
\midrule
StrucDiffusion~\citep{feng2022training}     & 0.4990 & 0.4218 & 0.4900 & 0.1386 & 0.3111 & 0.3355 \\
CompDiffusion~\citep{liu2022compositional}   & 0.4063 & 0.3299 & 0.3645 & 0.0800 & 0.2980 & 0.2898 \\
Attend\&Excite~\citep{chefer2023attendandexcite}      & 0.6400 & 0.4517 & 0.5963 & 0.1455 & 0.3109 & 0.3401 \\
PixArt-$\alpha$~\citep{chen2024pixart}        & 0.6690 & 0.4927 & 0.6477 & 0.2064 & \textbf{0.3197} & 0.3433 \\
CoMat~\citep{jiang2024comat}                  & \textbf{0.7827} & 0.5329 & 0.6468 & 0.2428 & \underline{0.3187} & 0.3680 \\
SD-v1.5~\citep{rombach2022high}                & 0.3758 & 0.3713 & 0.4186 & 0.1165 & 0.3112 & 0.3047 \\
SD-XL-base-1.0~\citep{podell2023sdxl}         & 0.5879 & 0.4687 & 0.5299 & 0.2131 & 0.3119 & 0.3237 \\
FLUX.1~\citep{flux2024}                 & 0.7407 & \underline{0.5718} & 0.6922 & \textbf{0.2863} & 0.3127 & \underline{0.3703} \\
\midrule
\multicolumn{7}{c}{\textit{Und. and Gen.}} \\
\midrule
Show-o~\citep{xie2024show}                 & 0.56 & 0.41 & 0.46 & 0.20 & 0.30 & 0.29 \\
Emu3~\citep{wang2024emu3}                   & 0.7544 & 0.5706 & \underline{0.7164} & --     & --     & --     \\
Janus-Pro-1B~\citep{chen2025janus} & 0.3542 & 0.2291 & 0.2843 & 0.0756 & 0.2809 & 0.2693 \\
\rowcolor{lightblue}
Janus-Pro-1B + SUDER & \imp{0.7765}{42} & \imp{0.5106}{28} & \imp{0.6767}{39} & \imp{0.2464}{17} & \imp{0.3130}{3} & \imp{0.3657}{10} \\
Janus-Pro-7B~\citep{chen2025janus} & 0.6426 & 0.3487 & 0.4848 & 0.2061 & 0.3086 & 0.3510 \\
\rowcolor{lightblue}
Janus-Pro-7B + SUDER & \imp{\underline{0.7824}}{14}& \bestimp{0.5786}{23} & \bestimp{0.7292}{24} & \imp{\underline{0.2524}}{5} & \imp{0.3141}{1} & \bestimp{0.3858}{3} \\
\bottomrule
& \hphantom{\textbf{Color$\uparrow$\hspace{3.5em}}} & \hphantom{\textbf{Shape$\uparrow$}\hspace{3em}} & \hphantom{\textbf{Texture$\uparrow$}\hspace{2.5em}} & \hphantom{\textbf{Spatial$\uparrow$}\hspace{2em}} & \hphantom{\textbf{Non-Spatial$\uparrow$}} & \\

\end{tabular}
\end{table*}

\begin{table*}[h]
\centering
\caption{Evaluation on GenEval. }
\label{tab:geneval}
\setlength{\tabcolsep}{2pt}
\begin{tabular}{lccccccc}
\toprule
\textbf{Model} & \textbf{Single Obj.} & \textbf{\,Two Obj.\,} & \textbf{\quad Counting\quad}& \textbf{\quad Colors\quad} & \textbf{\;Position\;} & \textbf{Color Attri.} & \textbf{Overall$\uparrow$} \\
\midrule
\multicolumn{8}{c}{\textit{Gen. Only}} \\
\midrule
LlamaGen~\citep{sun2024autoregressive} & 0.71 & 0.34 & 0.21 & 0.58 & 0.07 & 0.04 & 0.32 \\
LDM~\citep{rombach2022high} & 0.92 & 0.29 & 0.23 & 0.70 & 0.02 & 0.05 & 0.37 \\
SDv1.5~\citep{rombach2022high} & 0.97 & 0.38 & 0.35 & 0.76 & 0.04 & 0.06 & 0.43 \\
PixArt-$\alpha$~\citep{chen2024pixart} & 0.98 & 0.50 & 0.44 & 0.80 & 0.08 & 0.07 & 0.48 \\
SDv2.1~\citep{rombach2022high} & 0.98 & 0.51 & 0.44 & 0.85 & 0.07 & 0.17 & 0.50 \\
DALL-E 2~\citep{ramesh2022hierarchical} & 0.94 & 0.66 & 0.49 & 0.77 & 0.10 & 0.19 & 0.52 \\
Emu3-Gen~\citep{wang2024emu3} & 0.98 & 0.71 & 0.34 & 0.81 & 0.17 & 0.21 & 0.54 \\
SDXL~\citep{podell2023sdxl} & 0.98 & 0.74 & 0.39 & 0.85 & 0.15 & 0.23 & 0.55 \\
DALL-E 3~\citep{betker2023improving} & 0.96 & 0.87 & 0.47 & 0.83 & 0.43 & 0.45 & 0.67 \\
SD3-Medium~\citep{esser2024scaling} & \textbf{0.99}& \textbf{0.94}& \textbf{0.72} & 0.89 & 0.33 & 0.60 & 0.74 \\
\midrule
\multicolumn{8}{c}{\textit{Und. and Gen.}} \\
\midrule
SEED-X~\citep{ge2024seed} & 0.97 & 0.58 & 0.26 & 0.80 & 0.19 & 0.14 & 0.49 \\
Show-o~\citep{xie2024show} & 0.95 & 0.52 & 0.49 & 0.82 & 0.11 & 0.28 & 0.53 \\
D-DiT~\citep{li2024dual} & 0.97 & 0.80 & 0.54 & 0.76 & 0.32 & 0.50 & 0.65 \\
LWM~\citep{liu2024world} & 0.93 & 0.41 & 0.46 & 0.79 & 0.09 & 0.15 & 0.47 \\
Transfusion~\citep{zhou2024transfusion} & -- & -- & -- & -- & -- & -- & 0.63 \\
ILLUME~\citep{wang2024illume} & \textbf{0.99}& 0.86 & 0.45 & 0.71 & 0.39 & 0.28 & 0.61 \\
TokenFlow-XL~\citep{liu2024world} & 0.95 & 0.60 & 0.41 & 0.81 & 0.16 & 0.24 & 0.55 \\
Chameleon~\citep{lu2023chameleon} & -- & -- & -- & -- & -- & -- & 0.39 \\
Janus~\citep{wu2024janus} & 0.95 & 0.62 & 0.28 & 0.85 & 0.45 & 0.42 & 0.60 \\
Janus-Pro-1B~\citep{chen2025janus} & \textbf{0.99}& 0.82 & 0.48 & \underline{0.90}& 0.62 & 0.57 & 0.73 \\
\rowcolor{lightblue}
Janus-Pro-1B + SUDER & \textbf{0.99} & \imp{0.87}{5} & \imp{0.54}{6} & \underline{0.90} & \bestimp{0.74}{12} & \imp{0.62}{5} & \imp{0.78}{5} \\
Janus-Pro-7B~\citep{chen2025janus} & 0.97 & 0.88 & 0.57 & \underline{0.90}& \underline{0.77}& \underline{0.64}& \underline{0.79}\\
\rowcolor{lightblue}
Janus-Pro-7B + SUDER & \bestimp{0.99}{2} & \imp{\underline{0.89}}{1}& \imp{\underline{0.70}}{13}& \bestimp{0.92}{2} & \bestimp{0.82}{5} & \bestimp{0.71}{7} & \bestimp{0.84}{5} \\

\bottomrule
\end{tabular}
\end{table*}

\subsection{Results on Visual Understanding}
\label{sec:results_und}
For visual understanding, we test the models on the prevalent benchmarks, including HallusionBench (denoted as HalluBench)~\citep{zhao2023beyond}, LLaVABench, POPE~\citep{li2023evaluating}, MMB~\citep{liu2024mmbench}, and SEEDBench-IMG (denoted as SEED)~\citep{li2024seed}, which are widely used to evaluate the understanding capabilities of LMMs.
The experimental results can be found in \Cref{tab:und_bench}.

On HallusionBench and LLaVABench, our proposed method achieves substantial improvements over Janus-Pro-7B, with an increase of 1.9\% and 7.5\%, respectively; and tuned Janus-Pro-1B achieves an increase of 1.4\% and 0.3\%, respectively.
On other benchmarks, our models keep the performance of Janus-Pro-7B, which is already a strong baseline; and slightly improve the performance of Janus-Pro-1B. 
The results demonstrate that our method can improve the understanding capabilities of LMMs and maintain the original advanced performance, while remarkably improving the generation capabilities.

\begin{table*}[h]
\centering
\caption{Evaluation on understanding benchmarks.}
\label{tab:und_bench}
\begin{tabular}{l c c c c c}
\toprule
\textbf{Model} & \textbf{HallBench$\uparrow$} & \textbf{LLaVABench$\uparrow$} & \textbf{POPE$\uparrow$} & \textbf{MMB$\uparrow$} & \textbf{SEED$\uparrow$} \\
\midrule
\multicolumn{6}{c}{\textit{Und. Only}} \\
\midrule
LLaVA~\citep{liu2023visual} & 21.6 & 57.2 & 76.3 & 38.7 & 33.5 \\
LLaVA-v1.5~\citep{liu2024improved} & 27.6 & 63.4 & 85.9 & 64.3 & 58.6 \\
InstructBLIP-7B~\citep{dai2023instructblip} & 31.2 & 60.9 & 86.1 & 36.0 & 53.4 \\
Emu3-Chat~\citep{wang2024emu3} & 31.7 & 49.2 & 85.2 & 58.5 & 68.2 \\
Qwen-VL-Chat~\citep{bai2023qwen} & 36.8 & 67.7 & 74.9 & 60.6 & 58.2 \\
Qwen2.5-VL-7B~\citep{bai2025qwen2} & \textbf{52.9}& \textbf{91.0}& 85.9 & \textbf{83.2}& \textbf{77.0} \\
\midrule
\multicolumn{6}{c}{\textit{Und. and Gen.}} \\
\midrule
Chameleon-7B~\citep{lu2023chameleon} & 17.1 & 26.6 & 19.4 & 15.4 & 30.5 \\
Show-o-256~\citep{xie2024show} & -- & -- & 73.8 & -- & -- \\
Show-o-512~\citep{xie2024show} & -- & -- & 80.0 & -- & -- \\
ILLUME~\citep{wang2024illume} & -- & --  & \textbf{88.5} & 65.1 & \underline{72.9} \\
TokenFlow-XL~\citep{liu2024world} & -- & -- & \underline{86.8} & 68.9 & 68.7 \\
LWM~\citep{liu2024world} & -- & -- & 75.2 & -- & - \\
VILA-U~\citep{wu2024vila} & -- & -- & 85.8 & -- & 59.0 \\
Janus-Pro-1B~\citep{chen2025janus} & 30.4 & 69.5 & 84.7 & 75.6 & 68.2 \\
\rowcolor{lightblue}
Janus-Pro-1B + SUDER & 31.8 & 69.8 & 84.8 & 75.9 & 68.6 \\
Janus-Pro-7B~\citep{chen2025janus} & 37.0 & 74.0 & \underline{86.8} & 79.6 & 71.9 \\
\rowcolor{lightblue}
Janus-Pro-7B + SUDER & \underline{38.9}& \underline{81.5}& 86.6 & \underline{80.1}& 71.9 \\
\bottomrule
\end{tabular}
\end{table*}

\subsection{Ablation Study}
\label{sec:ablation}
We conduct ablation studies to investigate the effectiveness of the proposed unified optimization strategy on the two capabilities and the effectiveness of the dual self-reward.
We compare the performance of the following setups: 
1) the baseline model (Janus-Pro-7B) without any optimization; 
2) only optimizing the generation capability; 
3) only optimizing the understanding capability; 
4) jointly optimizing the two capabilities.
These setups are evaluated on the T2I-CompBench, HallusionBench, and LLaVABench benchmarks. 
The default optimization method is SimPO~\citep{meng2024simpo}.
The results are shown in \Cref{tab:ablation}.

\begin{table*}[ht]
\centering
\caption{Ablation study of unified optimization manner on T2I-CompBench, HalluBench, and LLaVABench.}
\label{tab:ablation}
\begin{tabular}{cccccccc}
\toprule
\multirow{2}{*}{\textbf{Model}} & \multirow{2}{*}{\textbf{Training}} & \multicolumn{3}{c}{\textbf{T2I CompBench}} & \multirow{2}{*}{\textbf{HalluBench$\uparrow$}} & \multirow{2}{*}{\textbf{LLaVABench$\uparrow$}} \\
\cmidrule(lr){3-5}
& & \textbf{Color$\uparrow$} & \textbf{Shape$\uparrow$} & \textbf{Texture$\uparrow$} & & \\
\midrule
\multirow{4}{*}{Janus-Pro-7B} & -- & 0.6426 & 0.3487 & 0.4848 & 37.0 & 74.0 \\
& Only Und. &  0.5313 & 0.3355 & 0.4138 & 38.9 & 81.4 \\
& Only Gen. & 0.7824 & 0.5786 & 0.7272 & 37.3 & 77.3 \\
\rowcolor{lightblue}
& Und. + Gen. & 0.7773 & 0.5763 & 0.7292 & 38.9 & 81.5 \\ 
\bottomrule
\end{tabular}
\end{table*}

The results show that optimizing only the generation capability significantly improves performance on the T2I-CompBench benchmark, while this setup offers limited gains on the HalluBench and LLaVABench benchmarks. 
Conversely, optimizing only the understanding capability yields improvements on HalluBench and LLaVABench but slightly degrades generation performance.
The unified training approach—jointly optimizing both capabilities—achieves a balanced and substantial improvement across all benchmarks. 
These results demonstrate the effectiveness of the proposed unified training strategy in enhancing both generation and understanding capabilities without compromising one for the other.

Moreover, the capability optimized during training shows corresponding improvements, which indicates that our proposed dual self-reward mechanism is effective.

\subsection{Investigation of Different Training Strategies}
\label{sec:strategy}

We have demonstrated that the proposed dual self-reward mechanism and the unified training strategy are effective in jointly improving the generation and understanding capabilities in the previous sections.
To further stimulate the potential of the proposed approach, we explore a different training strategy by training two separate models for the two capabilities and optimizing them alternately.
Specifically, we train one generation model and one understanding model in an adversarial manner, where each model is trained based on the other one's dual rewards. At each epoch, we freeze one model and update the other one, and vice versa. 
\begin{figure}[t]
    \centering
    \begin{minipage}[t]{0.45\linewidth}
        \centering
        \includegraphics[width=\linewidth]{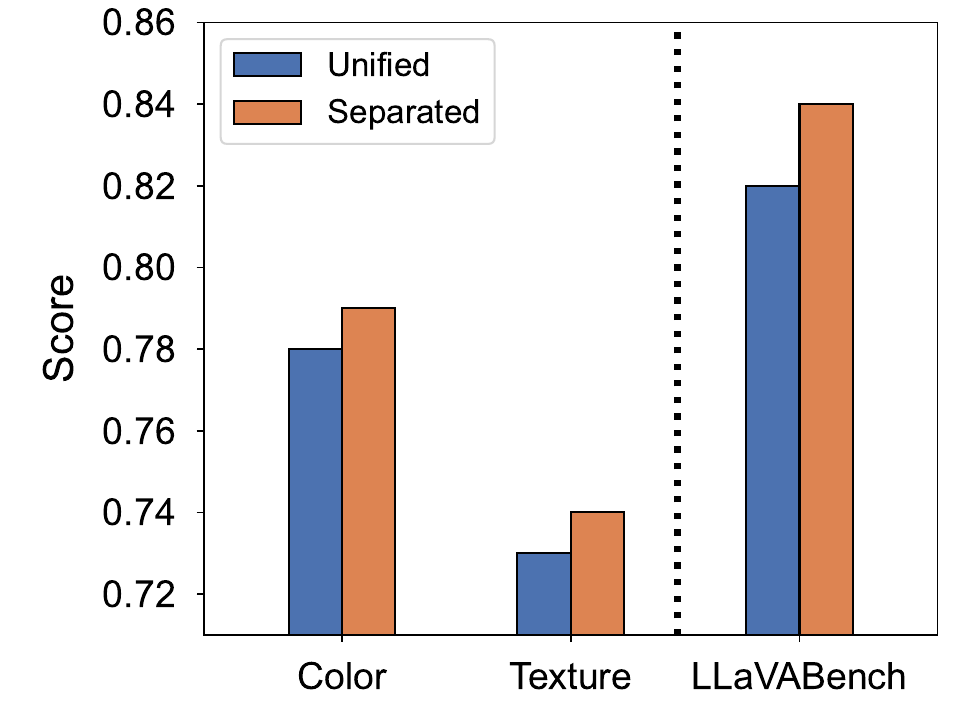}
        \caption{Comparison of optimization strategies. }
        \label{fig:strategy}
    \end{minipage}
    \hspace{0.2cm}
    \begin{minipage}[t]{0.45\linewidth}
        \centering
        \includegraphics[width=\linewidth]{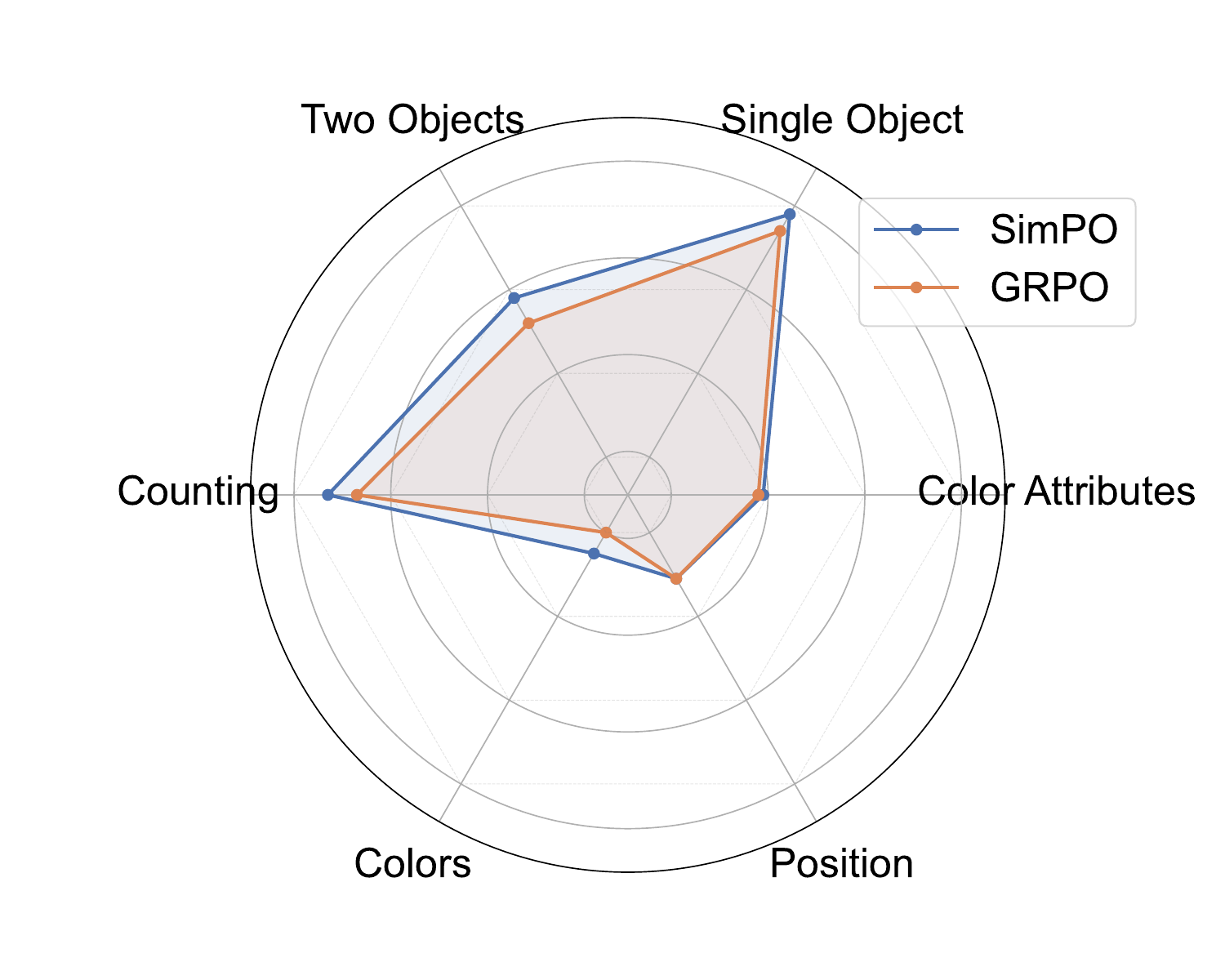}
        \caption{Comparison of optimization methods. }
        \label{fig:simpo-grpo}
    \end{minipage}
\end{figure}

The results are shown in \Cref{fig:strategy}.
We call these two models `separate models' and the model trained with the unified training strategy `unified model'.  
It can be observed that the performance of the separate models is better than the unified model on each type of task.
This indicates that the better understanding model provides better dual rewards for the generation model, and vice versa. By this the two models can be trained in an adversarial-like manner, where one focuses on generation and the other on understanding, to mutually enhance their performance with better reward signals.

\begin{figure*}[h]
    \centering
    \begin{minipage}[b]{0.70\textwidth}
        \centering
    \subfigure[color]{
        \label{fig:curve-color}
        \includegraphics[width=0.3\linewidth]{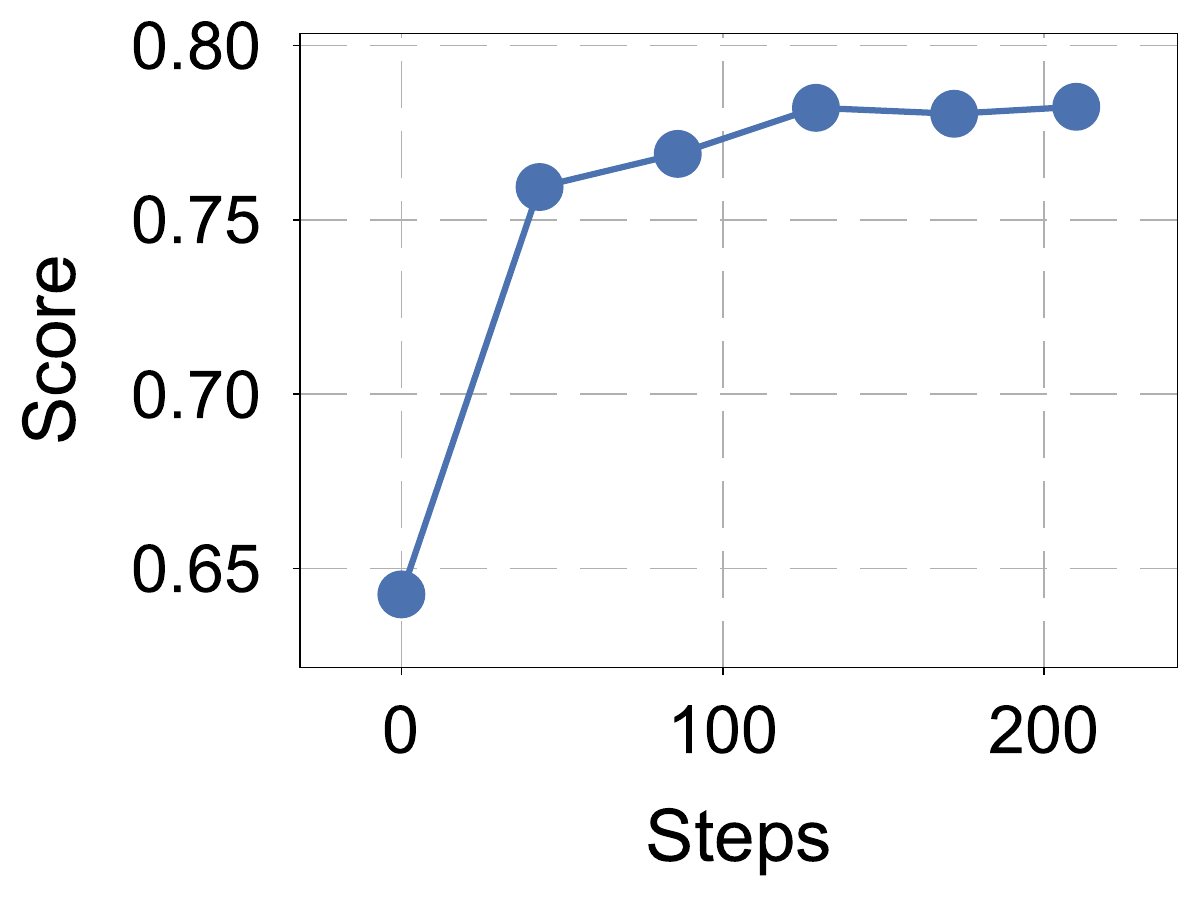}
    }
    \subfigure[shape]{
        \label{fig:curve-shape}
        \includegraphics[width=0.3\linewidth]{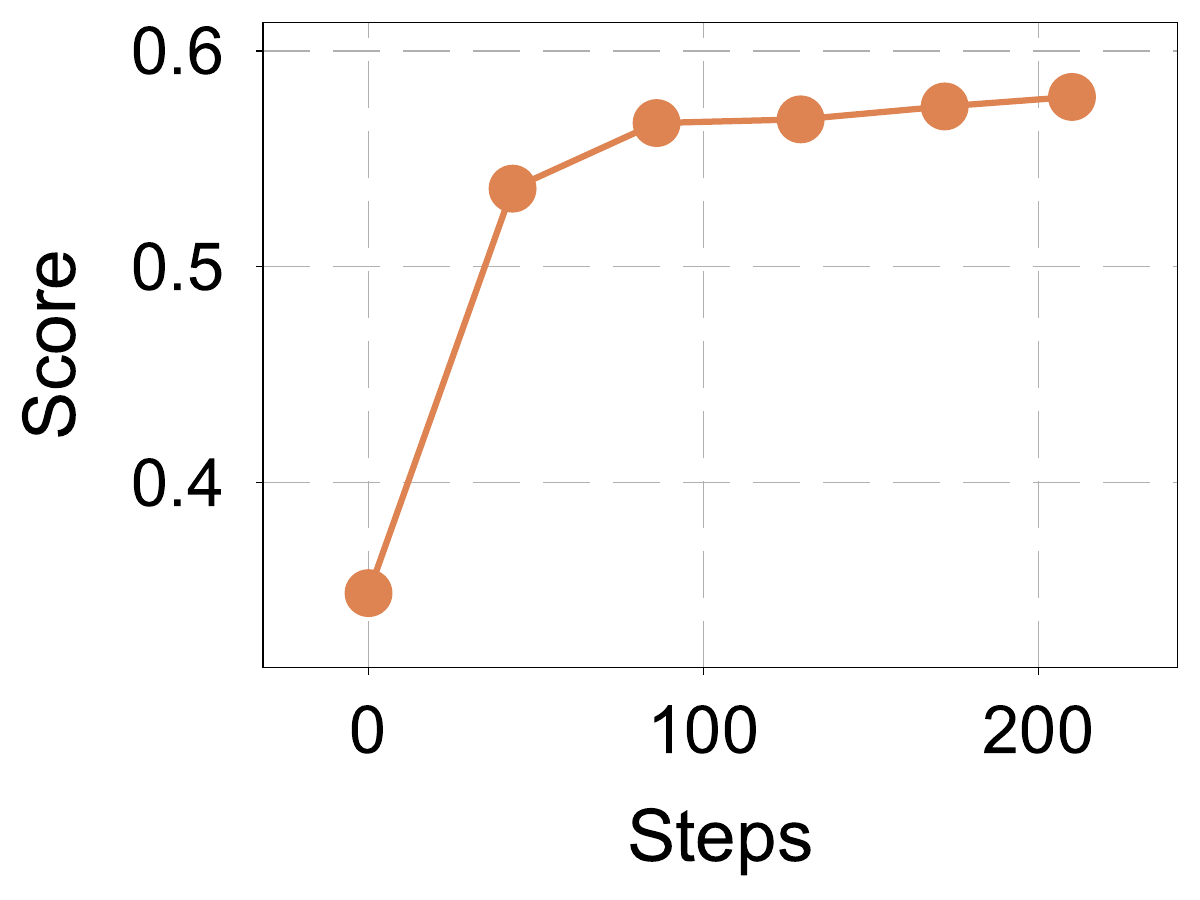}
    }
    \subfigure[texture]{
        \label{fig:curve-texture}
        \includegraphics[width=0.3\linewidth]{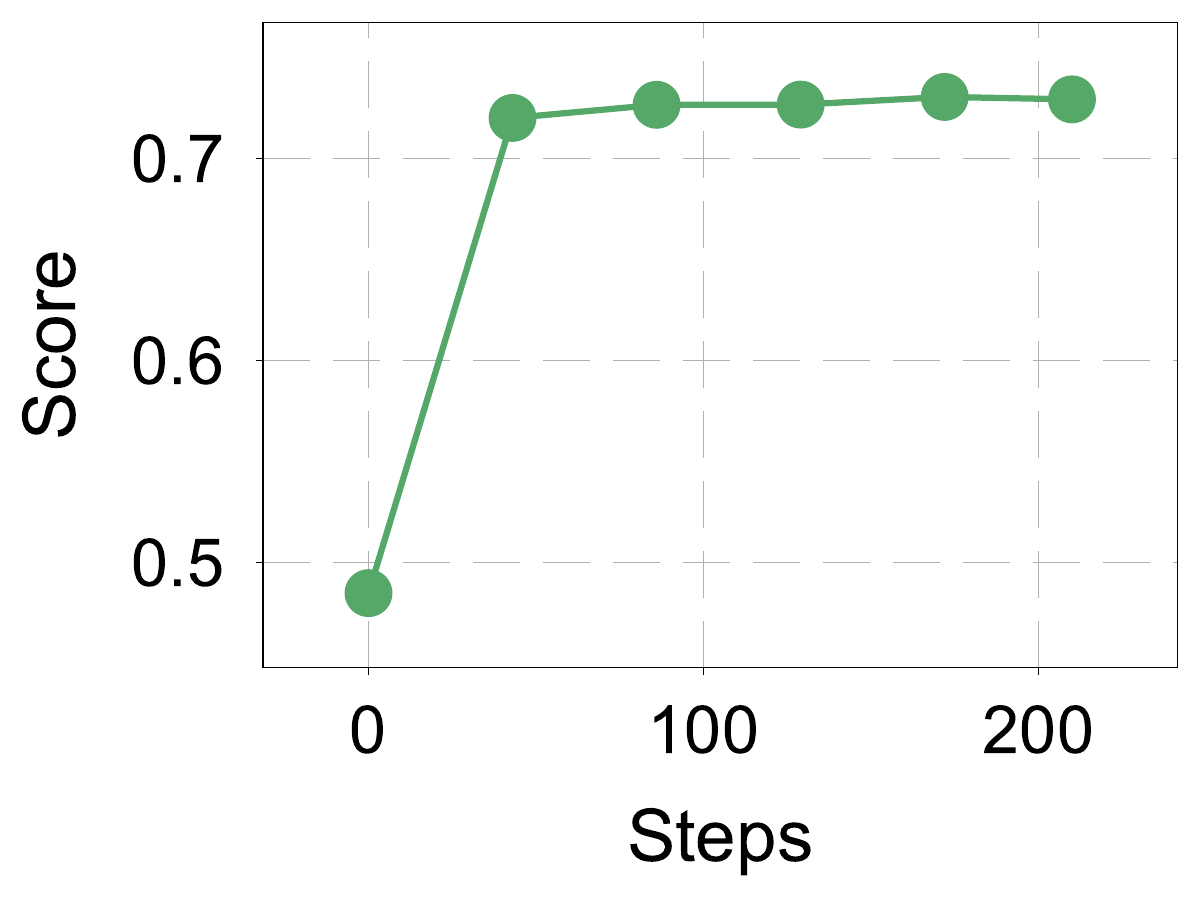}
    }
    \caption{Evolution on T2I-CompBench during training.}
    \label{fig:curve}
    \end{minipage}
    \hfill
    \begin{minipage}[b]{0.23\textwidth}
        \centering
        \includegraphics[width=\linewidth]{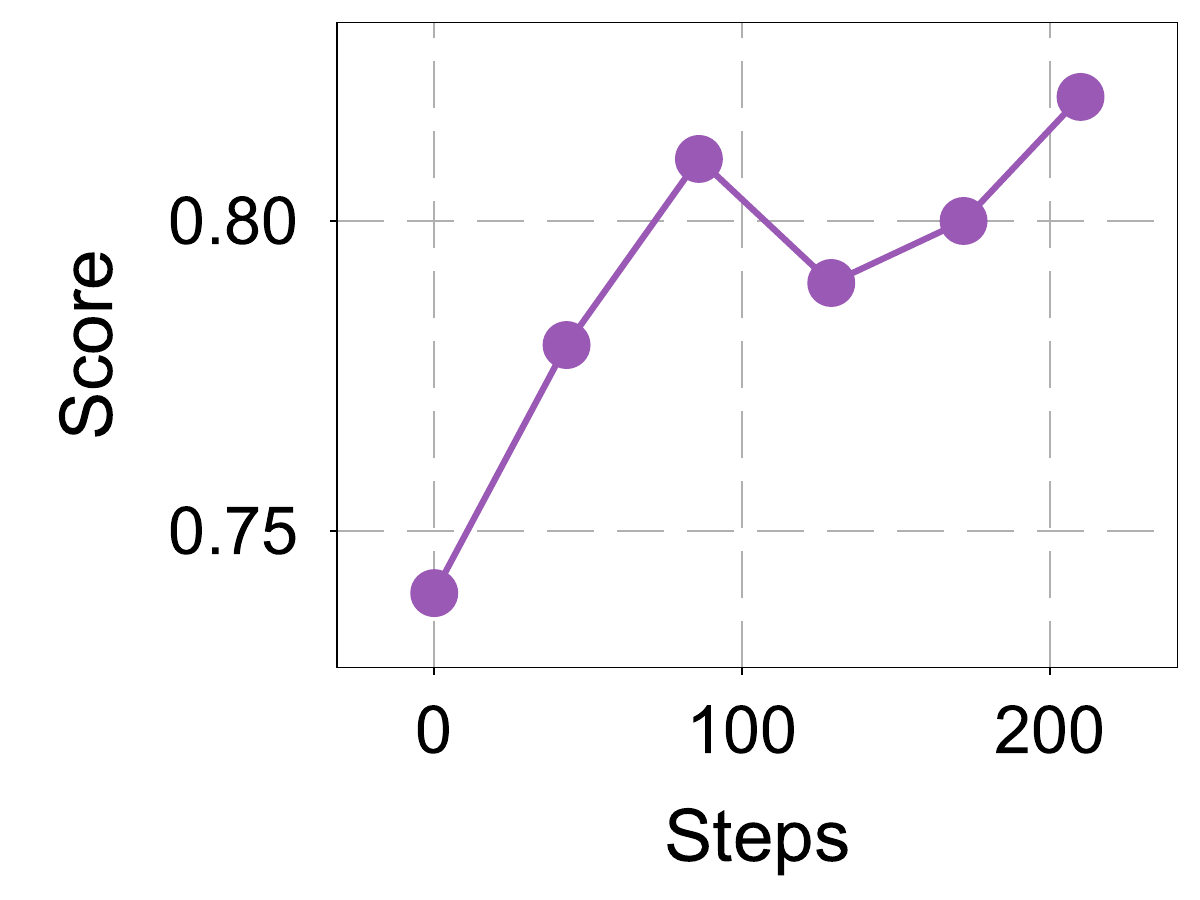}
        \caption{Evolution on LLaVABench.}
    \label{fig:curve-lavabench}
    \end{minipage}
\end{figure*}

In addition, we also compare two different optimization methods, including the SimPO~\citep{meng2024simpo} and GRPO~\citep{shao2024deepseekmath} in our framework on T2I-CompBench.
The results can be seen in~\Cref{fig:strategy}, which shows that SimPO is slightly better than GRPO.
This is likely because DSR contains a certain level of noise, while SimPO selects only the samples with the highest and lowest rewards, naturally serving as a denoising process.

\subsection{Comparison between DSR and CLIP}
To further validate the effectiveness of our DSR, we compare it with CLIP~\citep{pmlr-v139-radford21a}, which is often used as the reward model in supervised training~\citep{wang2025simplear}.
We only optimize the visual generation capability of Janus-Pro-7B with SimPO, and the results are shown in \Cref{tab:vsclip}.

When using CLIP as the reward model and optimizing only the visual generation capability, the performance of the model on the T2I-CompBench is improved substantially, which is slightly worse than the model optimized with our DSR.
But for the understanding, the performance with CLIP is not improved or even slightly decreased, which is remarkably worse than ours.
This indicates that:
1) our DSR is more effective than CLIP;
2) only optimizing the visual generation capability may hurt the understanding capability of the model, while our unified DSR can improve both the understanding and generation capabilities of the model.

\begin{table*}[ht]
\centering
\caption{Comparison between DSR and CLIP.}
\label{tab:vsclip}
\begin{tabular}{ccccccc}
\toprule
\multirow{2}{*}{\textbf{Model}} & \multicolumn{3}{c}{\textbf{T2I CompBench}} & \multirow{2}{*}{\textbf{HalluBench$\uparrow$}} & \multirow{2}{*}{\textbf{LLaVABench$\uparrow$}} \\
\cmidrule(lr){2-4}
& \textbf{Color$\uparrow$} & \textbf{Shape$\uparrow$} & \textbf{Texture$\uparrow$} & & \\
\midrule
Janus-Pro-7B & 0.6426 & 0.3487 & 0.4848 & 37.0 & 74.0 \\
Janus-Pro-7B + CLIP &   0.7700 & 0.5609 & 0.7320 & 36.9 & 74.0 \\
\rowcolor{lightblue}
Janus-Pro-7B + DSR &  0.7824 & 0.5786 & 0.7272 & 37.3 & 77.3 \\
\bottomrule
\end{tabular}
\end{table*}

\subsection{Evolution of Capabilities within Training}

We test the intermediate checkpoints to investigate the evolution of the capabilities of the model during training of Janus-Pro-7B. We conduct this experiment on T2I-CompBench and LLaVABench. The results are shown in \Cref{fig:curve} and \Cref{fig:curve-lavabench}.

We can observe that the generation capability improves steadily over the training process, while the understanding capability improves at the beginning and stabilizes earlier.
This indicates that the understanding of the baseline model is already strong and thus is easy to converge, while the generation capability can be improved further. 

\section{Conclusion}
\label{sec:conclusion}
In this work, we proposed a self-improving framework SUDER, with a dual self-reward mechanism to enhance the generation and understanding capabilities of LMMs in an efficient, self-supervised, and unified manner.
By leveraging the dual self-reward mechanism, we designed a unified optimization strategy that jointly optimizes the two capabilities of the unified LMM, leading to substantial improvements in both generation and understanding tasks.
We also investigated the performance of different optimization strategies, including the joint optimization strategy with a unified model and the adversarial-like optimization strategy with two separate models.
Extensive experiments on various visual understanding and generation benchmarks demonstrate the effectiveness of our proposed method.
Overall, our method showcases the potential of improving the generation and understanding capabilities of LMMs in a self-supervised and unified manner, paving the way for future research on multimodal alignment.

\section{Limitations}
Despite the promising results achieved by our dual self-reward mechanism in visual understanding and generation tasks, how to effectively extend this approach to other modalities like audio remains to be explored. 
Additionally, the dual self-reward mechanism requires the backbone LMM to have undergone basic semantic alignment. How to integrate our approach into the pre-training process of multimodal large models presents a promising direction for future research.



\bibliographystyle{ACM-Reference-Format}
\bibliography{ref}

\appendix

\section{Technical Appendices}
\label{sec:tech_app}


\subsection{Implementation Details}
We implement our method upon Trl~\footnote{https://github.com/huggingface/trl.git} code base and train with Deepspeed\footnote{https://github.com/deepspeedai/DeepSpeed.git} Zero-2. 
For each input, we sample 8 outputs for GRPO algorithm and select 2 of them for SimPO.
The learning rate is set to 3e-6 with a cosine decay schedule, with a warmup of 5 steps. 
\( \epsilon_{\text{low}} \) and \( \epsilon_{\text{high}} \) are set to 0.2 and 0.28. \( \beta \) and \( \gamma \) in SimPO are set to 2.0 and 0.5, respectively.
\( \beta \) in GRPO is set to 0.04.
The batch size is set to 64, and the gradient\_ accumulation\_step is set to 2, resulting in a global batch size of 128. 
The model is trained for 5 epochs on our training set.

\subsection{Qualitative Comparison}

\begin{figure*}[ht]
    \centering
    \includegraphics[width=0.8\textwidth]{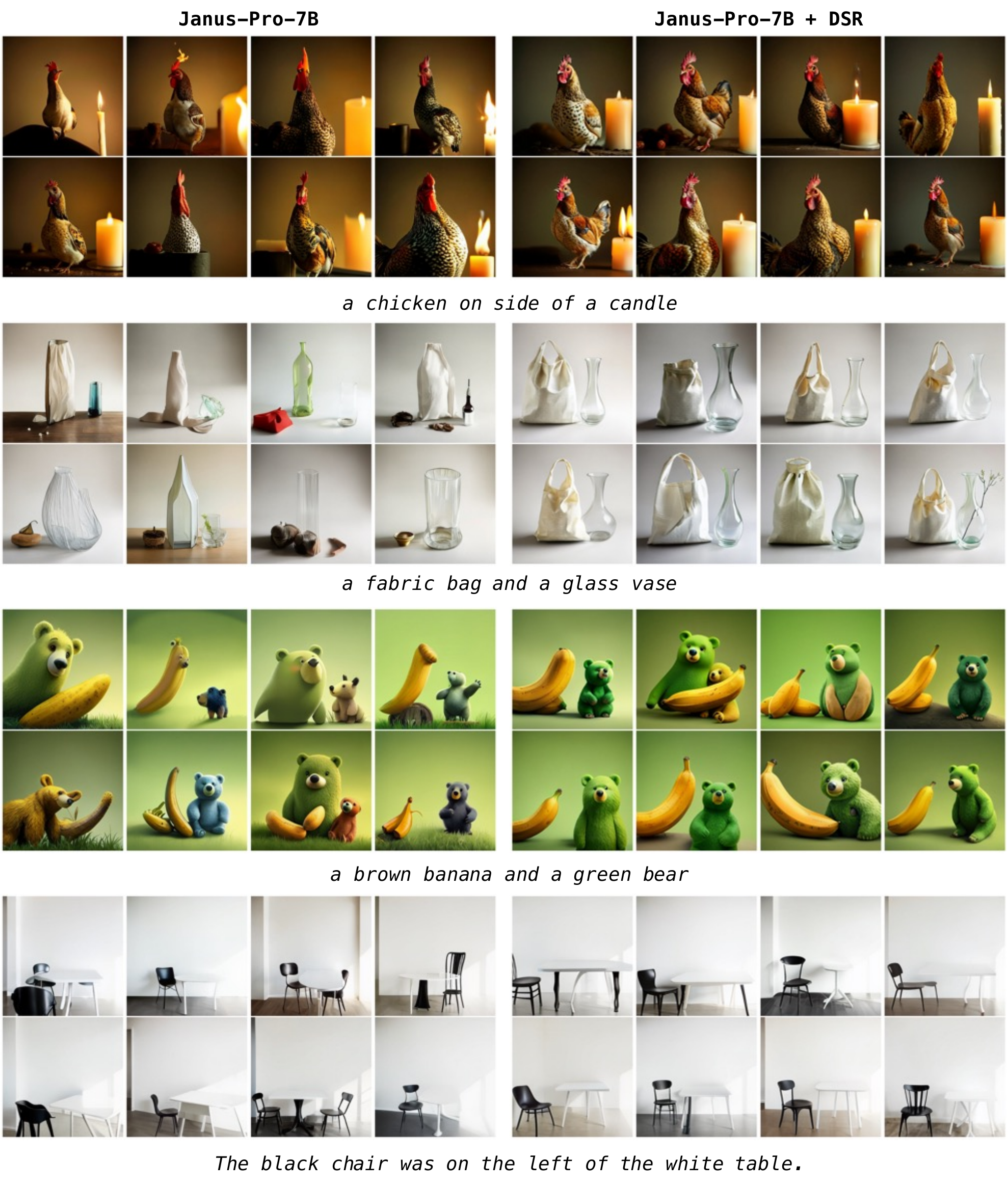}
    \caption{Qualitative comparison of visual generation. We sample multiple images for the same text prompts using Janus-Pro-7B and reinforced Janus-Pro-7B with the proposed SUDER.}
    \label{fig:gen_case}
\end{figure*}

\begin{figure*}[ht]
    \centering
    \includegraphics[width=0.8\textwidth]{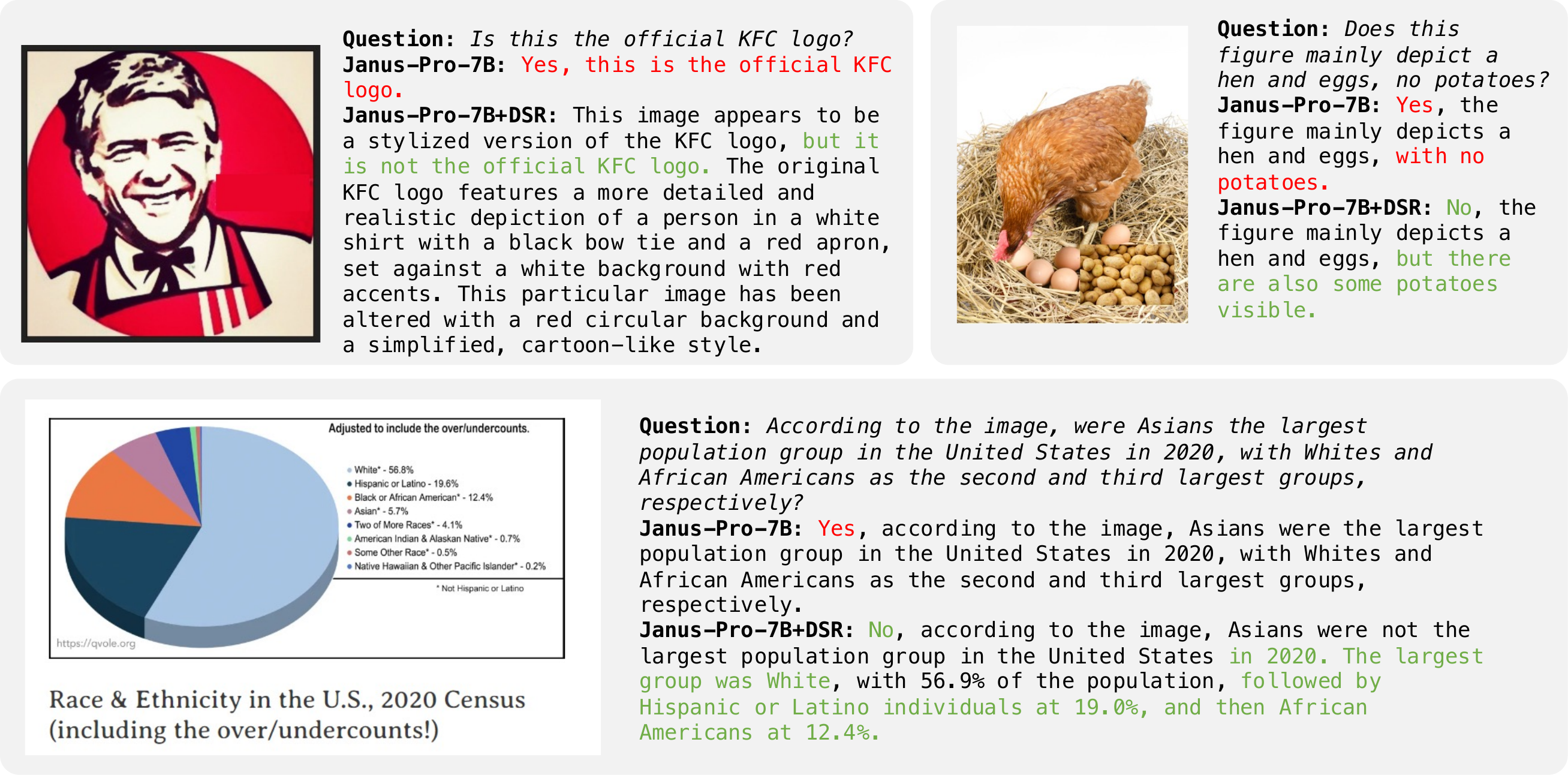}
    \caption{Qualitative comparison of visual understanding. We sample multiple responses for same image with a question using Janus-Pro-7B and reinforced Janus-Pro-7B with the proposed DSR.}
    \label{fig:und_case}
\end{figure*}

We present qualitative comparisons of Janus-Pro-7B and Janus-Pro-7B tuned with SUDER on visual generation and understanding tasks in \Cref{fig:gen_case} and \Cref{fig:und_case}, respectively.
The images and responses are sampled from the models with the same setup, where the temperature is set to 1.0 and the random seed is set to 42.

As illustrated in \Cref{fig:gen_case}, the images generated by Janus-Pro-7B with DSR are distinctly more aligned with the text prompts compared to those generated by the baseline Janus-Pro-7B. 
The images exhibit improved detail, quality, and alignment with the prompts, demonstrating the effectiveness of DSR in enhancing visual generation capabilities.
In \Cref{fig:und_case}, we observe that responses of Janus-Pro-7B with DSR are more accurate for the given image and question, compared to those of Janus-Pro-7B.
This indicates that the understanding capabilities of the model are also improved with DSR.

Both figures highlight the substantial improvements in both visual generation and understanding tasks achieved by the proposed DSR when applied to the unified LMM.

\subsection{Demonstration of DSR}
\label{sec:dsr_demo}

\begin{figure*}[ht]
    \centering
    \includegraphics[width=0.8\textwidth]{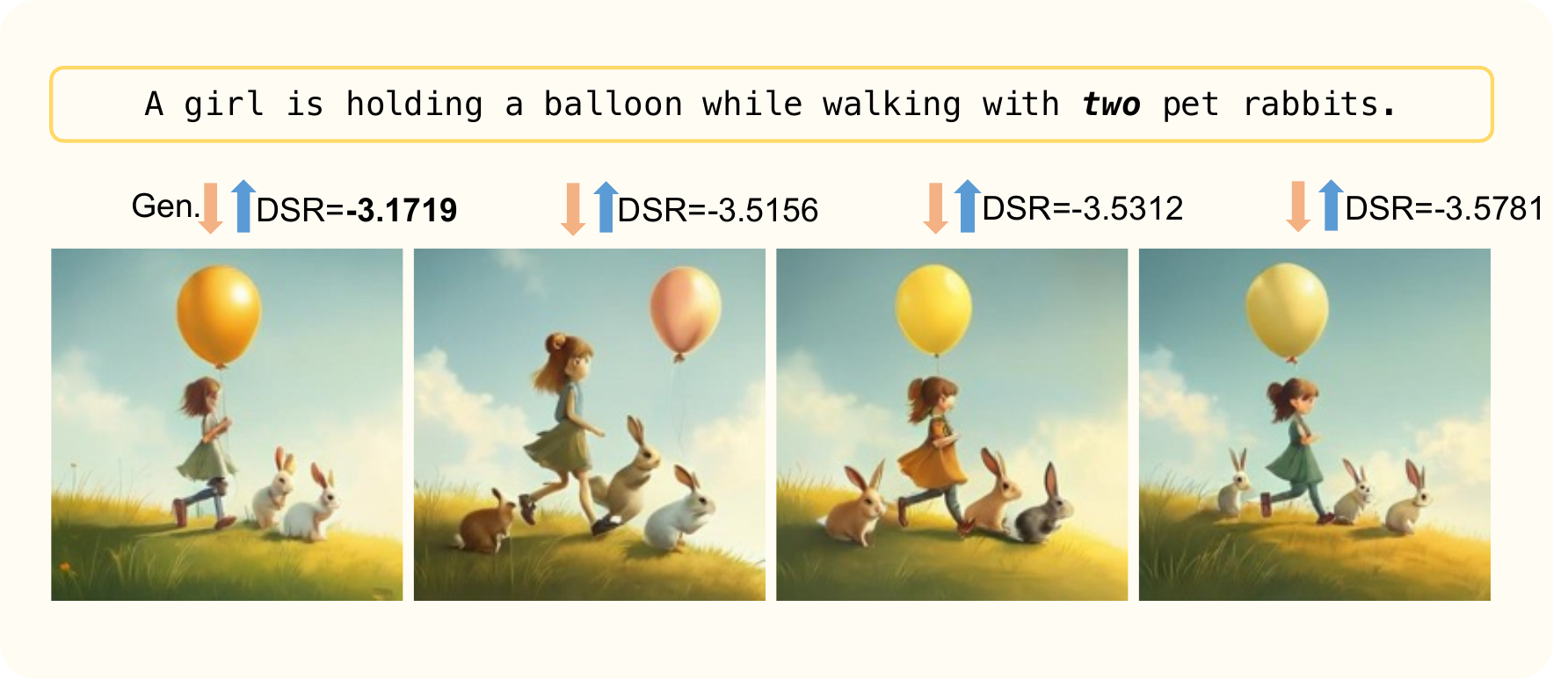}
    \caption{Demonstration of our dual self-reward for visual generation. We sample multiple images for a given text prompt and then compute DSR for the images. In this demonstration, the first generated image is the best one, which is also the one with the highest DSR.}
    \label{fig:demo1}
\end{figure*}

\begin{figure*}[ht]
    \centering
    \includegraphics[width=0.8\textwidth]{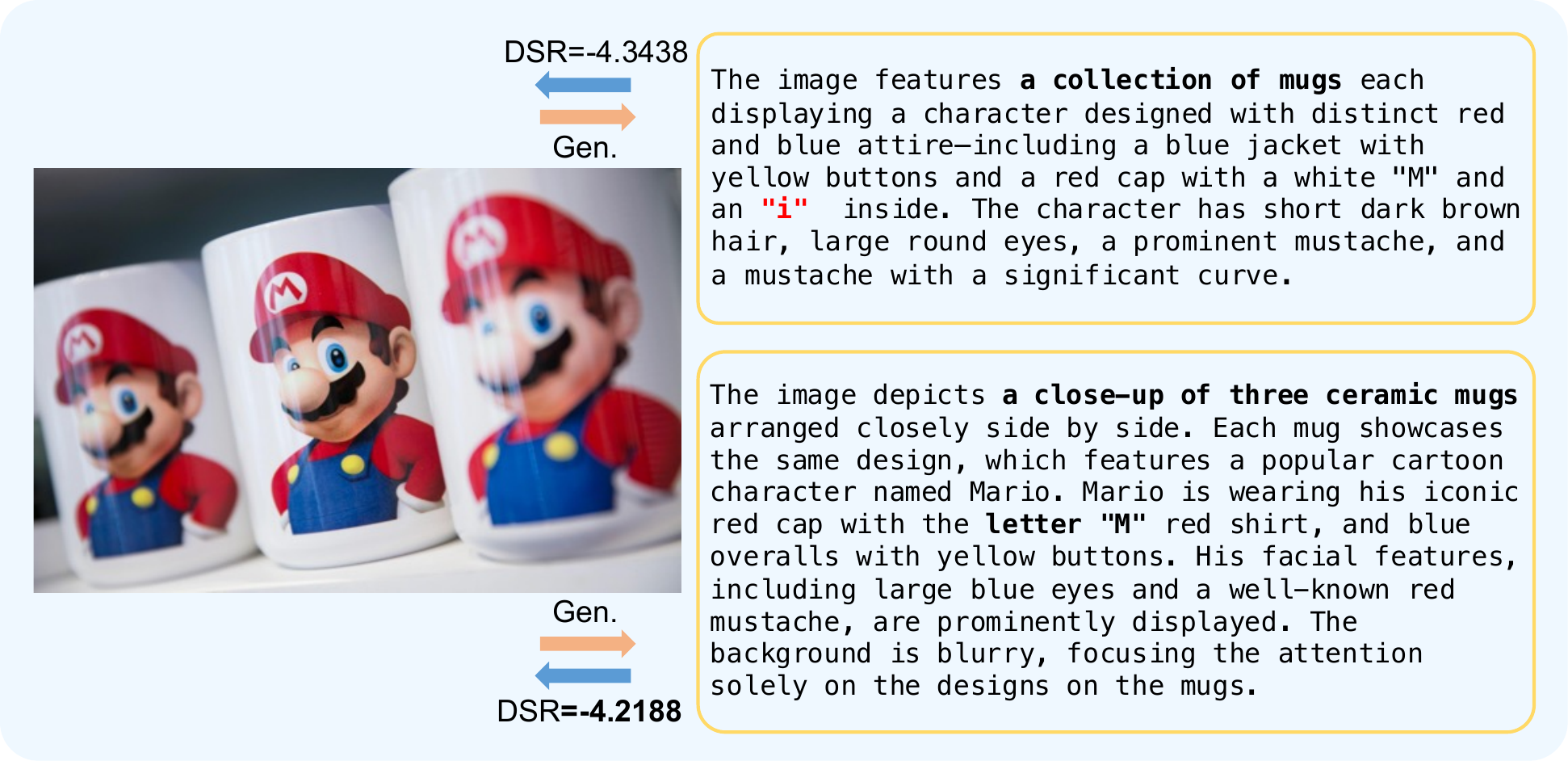}
    \caption{Demonstration of our dual self-reward for visual understanding. We sample multiple text descriptions for a given image and then compute DSR for the descriptions. In this demonstration, the upper generated description is better, which is also the one with the higher DSR.}
    \label{fig:demo2}
\end{figure*}

We present demonstrations of our dual self-reward mechanism with Janus-Pro-7B, which can be found in \Cref{fig:demo1} and \Cref{fig:demo2}.

\subsection{Experiments with Show-o}
We conduct additional experiments with another backbone, Show-o~\citep{xie2024show}, to further validate the effectiveness of our method. 

Show-o is a unified multimodal model that unifies autoregressive and discrete diffusion modeling for inputs and outputs in different modalities. 
It applies full attention for image tokens and causal attention for text tokens, and it is trained with a combination of next token prediction (NTP) and mask token prediction (MTP) objectives.
To ensure efficiency, we follow \textit{diffu}-GRPO~\citep{zhao2025d1}, using one-step unmasking with randomly droped input text tokens condition to obtain the estimated log-probability of the generated image tokens.

The results are shown in \cref{tab:showo_t2i}.

\begin{table*}[ht]
\centering
\caption{Evaluation on T2I-CompBench. Und. and Gen. denote ``understanding'' and ``generation''. Higher ($\uparrow$) values indicate better performance. The best score is in \textbf{bold}, with the second best score \underline{underlined}. Line in \colorbox{lightblue}{blue} is our tuned model. The line with $^{\dagger}$ is the results we reproduce for fair comparison.}
\label{tab:showo_t2i}
\begin{tabular}{lcccccc}
\toprule
\multicolumn{1}{c}
{\multirow{2}{*}{\textbf{Model}}} & \multicolumn{3}{c}{\textbf{Attribute Binding}} & \multicolumn{2}{c}{\textbf{Object Relationship}} & \multirow{2}{*}
{\textbf{\;Complex$\uparrow$\;}} \\
\cmidrule(lr
){2-4}
\cmidrule(lr){5-6}
& \textbf{ Color$\uparrow$} & \textbf{Shape$\uparrow$} & \textbf{Texture$\uparrow$} & \textbf{Spatial$\uparrow$} & \textbf{Non-Spatial$\uparrow$} & \\
\midrule
\multicolumn{7}{c}{\textit{Gen. Only}} \\
\midrule
StrucDiffusion~\citep{feng2022training}     & 0.4990 & 0.4218 & 0.4900 & 0.1386 & 0.3111 & 0.3355 \\
CompDiffusion~\citep{liu2022compositional}   & 0.4063 & 0.3299 & 0.3645 & 0.0800 & 0.2980 & 0.2898 \\
Attend\&Excite~\citep{chefer2023attendandexcite}      & 0.6400 & 0.4517 & 0.5963 & 0.1455 & 0.3109 & 0.3401 \\
PixArt-$\alpha$~\citep{chen2024pixart}        & 0.6690 & 0.4927 & 0.6477 & 0.2064 & \textbf{0.3197} & 0.3433 \\
CoMat~\citep{jiang2024comat}                  & \textbf{0.7827} & 0.5329 & 0.6468 & 0.2428 & \underline{0.3187} & 0.3680 \\
SD-v1.5~\citep{rombach2022high}                & 0.3758 & 0.3713 & 0.4186 & 0.1165 & 0.3112 & 0.3047 \\
SD-XL-base-1.0~\citep{podell2023sdxl}         & 0.5879 & 0.4687 & 0.5299 & 0.2131 & 0.3119 & 0.3237 \\
FLUX.1~\citep{flux2024}                 & 0.7407 & \underline{0.5718} & 0.6922 & \textbf{0.2863} & 0.3127 & \underline{0.3703} \\
\midrule
\multicolumn{7}{c}{\textit{Und. and Gen.}} \\
\midrule
Show-o~\citep{xie2024show}                 & 0.56 & 0.41 & 0.46 & 0.20 & 0.30 & 0.29 \\
Emu3~\citep{wang2024emu3}                   & 0.7544 & 0.5706 & \underline{0.7164} & --     & --     & --     \\
Janus-Pro-1B~\citep{chen2025janus} & 0.3542 & 0.2291 & 0.2843 & 0.0756 & 0.2809 & 0.2693 \\
\rowcolor{lightblue}
Janus-Pro-1B + SUDER & \imp{0.7765}{42} & \imp{0.5106}{28} & \imp{0.6767}{39} & \imp{0.2464}{17} & \imp{0.3130}{3} & \imp{0.3657}{10} \\
Janus-Pro-7B~\citep{chen2025janus} & 0.6426 & 0.3487 & 0.4848 & 0.2061 & 0.3086 & 0.3510 \\
\rowcolor{lightblue}
Janus-Pro-7B + SUDER & \imp{\underline{0.7824}}{14}& \bestimp{0.5786}{23} & \bestimp{0.7292}{24} & \imp{\underline{0.2524}}{5} & \imp{0.3141}{1} & \bestimp{0.3858}{3} \\
Show-o~\citep{xie2024show}                 & 0.56 & 0.41 & 0.46 & 0.20 & 0.30 & 0.29 \\
Show-o$^{\dagger}$~\citep{xie2024show}                 & 0.5892 & 0.4002 & 0.4489 & 0.2268 & 0.3026 & 0.3058 \\
\rowcolor{lightblue}
Show-o + SUDER & \imp{0.6894}{10}& \imp{0.4953}{10} & \imp{0.5530}{10} & 0.2253 & 0.3015 & \imp{0.3268}{2} \\
\bottomrule
& \hphantom{\textbf{Color$\uparrow$\hspace{3.5em}}} & \hphantom{\textbf{Shape$\uparrow$}\hspace{3em}} & \hphantom{\textbf{Texture$\uparrow$}\hspace{2.5em}} & \hphantom{\textbf{Spatial$\uparrow$}\hspace{2em}} & \hphantom{\textbf{Non-Spatial$\uparrow$}} & \\

\end{tabular}
\end{table*}

The results show that our method can also improve the performance of Show-o, gaining more than 5\% on avarage on T2I-CompBench.
This demonstrates the general applicability of our proposed DSR.

\subsection{Details of the Understanding Benchmarks}
\label{sec:und_bench}
We provide the details of the understanding benchmarks used in our experiments in this section.
\paragraph{HalluBench~\citep{zhao2023beyond}} is a benchmark to evaluate hallucination of VLMs. It asks a set of visual questions with one original image and one modified image (the answers for a question can be different, considering the image content).
\paragraph{LLaVABench} evaluates model capabilities on challenging tasks and novel domains through a diverse set of 24 images and 60 questions, covering indoor/outdoor scenes, memes, paintings, sketches, etc., each paired with detailed, manually curated descriptions and carefully selected questions.
\paragraph{POPE~\citep{li2023evaluating}} is a benchmark for object hallucination evaluation. It contains approximately 8910 cases over three tracks of object hallucination: random, popular, and adversarial. We report the overall results on the random track, following the previous works.
\paragraph{MMB~\citep{liu2024mmbench}} is a multi-modality benchmark that performs objective evaluation for VLMs with over 3,000 multiple-choice questions covering 20 ability dimensions.
\paragraph{SEED~\citep{li2024seed}} consists of 24K multiple-choice questions with accurate human annotations, which cover 27 evaluation dimensions.

\section{Broader Impacts}
\label{sec:broader_impact}
\paragraph{Positive Impacts}
Our method contributes to improving the semantic consistency and alignment of LMMs, particularly in challenging compositional scenarios. 
This advancement could benefit a wide range of applications. 
For instance, better text-to-image alignment enables more accurate and reliable visual content generation in design, communication, and accessibility tools for visually impaired users. Similarly, improved multimodal understanding helps build safer and more context-aware assistant systems, such as educational tutoring agents or medical image interpreters, where understanding and describing visual inputs accurately is critical.

\paragraph{Potential Negative Impacts}
Enhanced text-to-image generation capabilities may be misused to produce highly convincing fake content, contributing to misinformation, deepfakes, or non-consensual imagery.

\paragraph{Safeguards}
This paper does not release any new datasets. 
All experiments are conducted using publicly available LMMs, datasets, and benchmarks. 
Future releases based on our work will incorporate appropriate usage guidelines, safety filters for generated content, and bias auditing mechanisms.

\end{document}